\documentclass[10pt, a4paper]{article}
\usepackage{ltc23}
\usepackage{graphicx}
\usepackage{hyperref}
\usepackage{scalerel}
\usepackage{color,soul}
\usepackage{bm}
\usepackage{xcolor}
\usepackage{tcolorbox}
\usepackage[T1]{fontenc} 

\definecolor{submissioncolor}{RGB}{245,245,245} 
\newtcolorbox{submissionbox}{
  colback=submissioncolor,
  colframe=black,
  boxrule=1pt, 
  arc=3pt, 
  boxsep=0pt,
  left=4pt,
  right=4pt,
  top=4pt,
  bottom=4pt,
  fontupper=\small\itshape,
}

\title{XAI in Computational Linguistics: \\  Understanding Political Leanings in the Slovenian Parliament}

\name{Bojan Evkoski$^{\ast}$$^{\dagger}$, Senja Pollak$^{\ast}$} 
\address{ $^{\ast}$Jozef Stefan Institute \\
$^{\dagger}$ Jozef Stefan International Postgraduate School \\
                  bojan.evkoski@ijs.si  \\
                  senja.pollak@ijs.si \\}

\abstract{The work covers the development and explainability of machine learning models for predicting political leanings through parliamentary transcriptions. We concentrate on the Slovenian parliament and the heated debate on the European migrant crisis, with transcriptions from 2014 to 2020. We develop both classical machine learning and transformer language models to predict the left- or right-leaning of parliamentarians based on their given speeches on the topic of migrants. With both types of models showing great predictive success, we continue with explaining their decisions. Using explainability techniques, we identify keywords and phrases that have the strongest influence in predicting political leanings on the topic, with left-leaning parliamentarians using concepts such as \textit{people} and \textit{unity} and speak about \textit{refugees}, and right-leaning parliamentarians using concepts such as \textit{nationality} and focus more on \textit{illegal migrants}. This research is an example that understanding the reasoning behind predictions can not just be beneficial for AI engineers to improve their models, but it can also be helpful as a tool in the qualitative analysis steps in interdisciplinary research.}

\begin{document}

\maketitleabstract

\begin{submissionbox}
The final formatted version of this publication was published in Proceedings of the 10th Language and Technology Conference (LTC 2023), April 2023, and is available online at \href{https://dx.doi.org/10.14746/amup.9788323241775}{dx.doi.org/10.14746/amup.9788323241775}.
\end{submissionbox}

\section{Introduction}

Artificial intelligence, in particular machine learning, is extensively used for solving many real-world problems both in research~\cite{jordan2015machine} and industry~\cite{shinde2018review}. In recent years, with AI being more present and the introduction of stricter guidelines~\cite{interpol2019artificial} and regulations~\cite{european2020gdpr}, it became apparent that the sheer power of machine learning models for predicting tasks does not justify their widespread potentially inconsiderate usage. Today, understanding and explaining the previously considered black-box models is equally important as developing and training them. This process enables engineers to detect biases in the model, come up with ideas for improvement, and most importantly examine the security of the system. Explainable AI (or simply XAI) is gradually becoming a must both in industry and research.

Another aspect of AI interpretability is its effect of bridging the gap between heavy quantitative black-box research and qualitative research more common among humanities and social sciences scholars. As machine learning engineers get feedback from a model on why it makes a particular prediction, humanities and social sciences scholars would take these signals to further examinations that can lead to more qualitative explanations of why certain features play a significant role in a particular problem.

This work covers a case of interdisciplinary research between computational linguistics (or natural language processing) and political science where we explain AI models to gain insight from a political linguistics perspective. We focus on parliamentary debates, a salient research topic in both humanities and social science disciplines, such as sociology, political science, sociolinguistics, and history~\cite{skubic2022parliamentary}. Our goal is to bring more insight into the speeches of parliamentarians (MPs) of different political leanings. First, by training machine learning models to predict if a parliamentarian is left or right-leaning based on their speech. And then, by using explainability techniques on the models, extract data and derive knowledge on what actually differentiates left and right political speeches in the parliament. To make things more politically relevant and methodologically clear, we are focusing on the concerning topic of migrants and the European migration crisis~\cite{barlai2017migrant} where the left and right have evident divergent stances: left-leaning parties showing consistent support to immigrants from Asia and Africa throughout Europe, and right-leaning parties showing moderate to strong opposition against immigration to Europe and their country in particular~\cite{van2021conditional}. We apply our analysis to the Slovenian Parliament from 2014 to 2020. For data, we use the open-access ParlaMint dataset~\cite{erjavec2022parlamint} which provides complete parliamentary transcriptions for 17 countries, including Slovenia.

The paper is organized as follows. In Section~\ref{related_work}, we briefly cover the related work on using explainable AI in social sciences, as well as recent applications of computational linguistics for parliamentary debates. In Section~\ref{data} we provide a description of the ParlaMint SI dataset we are working with, as well as the preprocessing steps required before training the models. In Section~\ref{modeling} we dive into the training of the classical machine learning and the modern language models. Finally, we discuss the results in Section~\ref{results_discussion} and conclude in Section~\ref{conclusion}.

\section{Related Work}
\label{related_work}

By the end of the 2010s and the beginning of 2020s, explainable AI has become a topic that involved both computer and social scientists in discussions on how to interpret and use the knowledge drawn from model explanations~\cite{miller2019explanation}. Social scientists urge the importance of qualitative investigation experts joining in XAI projects~\cite{johs2022explainable}. Yet, social qualitative investigation done by non-experts is still very common. Combining qualitative and quantitative approaches (mixed research) has a variety of hardships~\cite{brannen2017combining}, and researchers from both sides of the spectrum rarely have insight into the benefits of combining knowledge with the opposite side. With this work focusing on a specific social theme, we show an example of how XAI research can be the initial step for a more thorough qualitative approach~\cite{molina2016mixed}.

Parliamentary discourse, although extensively researched in a qualitative setting~\cite{ilie2015parliamentary}, is still underexplored in a quantitative manner. \newcite{naderi2015argumentation} use computational methods to analyze framing structures in speeches of the Canadian Parliament. \newcite{greene2015unveiling} use dynamic topic modeling to explore the evolution of the political agenda in the European Parliament. \newcite{eskicsar2022emotions} analyze the emotion structure of speeches in the Turkish ParlaMint dataset.

Due to the European Migration crisis in 2015, researchers have been studying the political stances of politicians~\cite{wallaschek2020contested}, news media~\cite{krotofil2018between}, and the effect of social media on the integration of migrants~\cite{alencar2018refugee}. Computational techniques were also applied. For example, \cite{greussing2017shifting} use techniques such as the bag-of-words and principal component analysis to understand media framing on the topic, while  \cite{heidenreich2019media} apply topic modeling to observe the dynamics of the migration narrative. In a more recent work by \cite{skubic2022networks}, the authors create mention networks for multiple parliaments on the topic of migration and analyze the role of gender on speech influence.

\section{Data}
\label{data}

\begin{figure}[t!]
\centering
\includegraphics[width=0.45\textwidth]{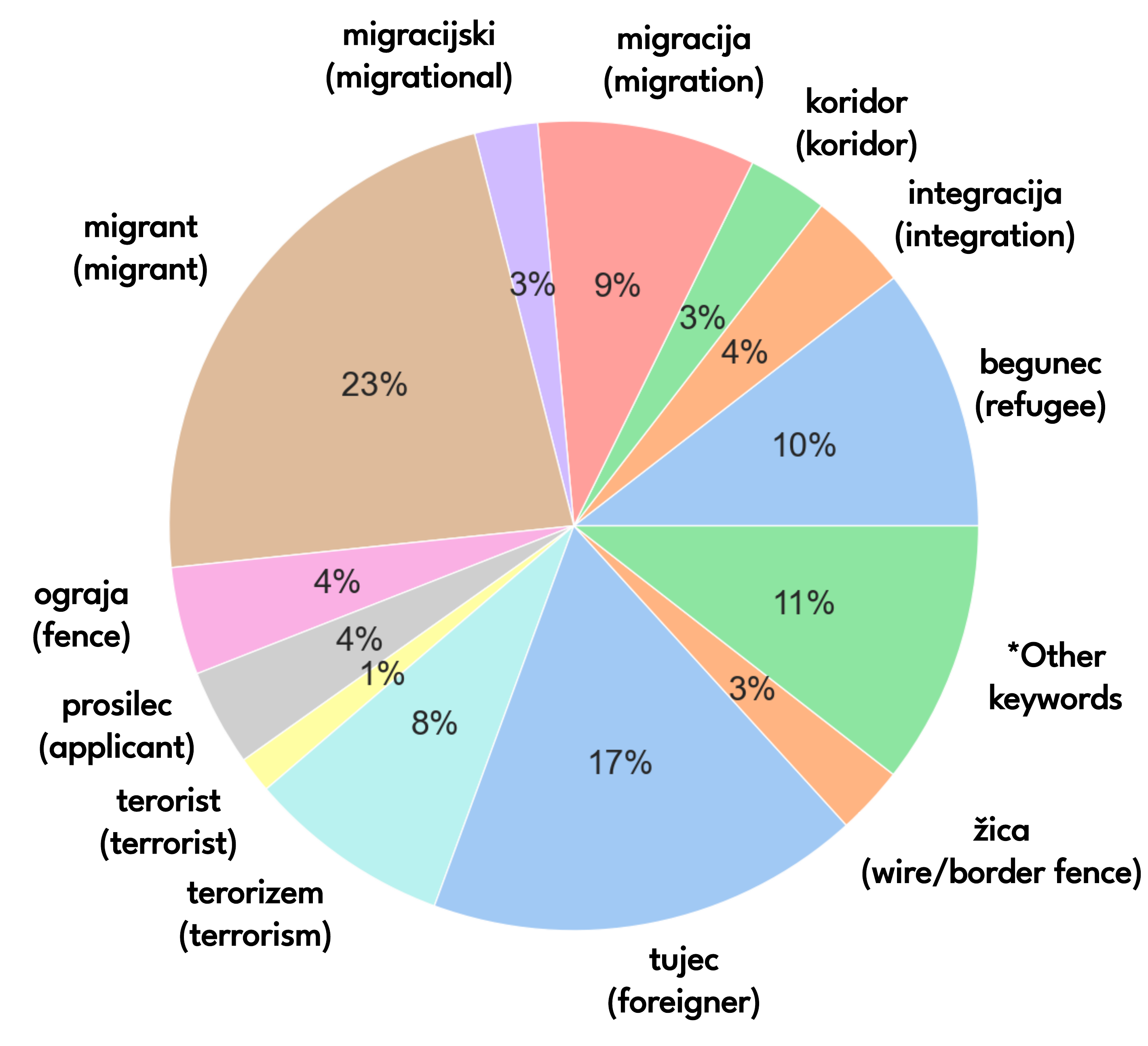}
\vspace*{-3ex}
\caption{\textbf{Pie chart on the migration keywords.}}
\vspace*{-1ex}
\label{figs:piechart}       
\end{figure}

Our choice of parliamentary transcriptions is the richly annotated ParlaMint subset of the Slovenian Parliament. It covers all sessions of the National Assembly from August 2014 to July 2020, with more than 20M words. In order to prepare the data for training a political leaning model based on the transcriptions, we first applied a few preprocessing steps according to the metadata. First, we removed all guest and Chair speeches, leaving only regular MPs. Next, used the party name data for each parliamentarian to extract the main label (right or left) using Wikipedia's "political position" English metadata on the party. Far-right, right, and center-right are labeled as right; far-left, left and center-left are labeled as left. Finally, we applied speech selection according to the topic of our interest --- migration. For this purpose, we prepared a set of keywords directly connected to the migration topic\footnote{The list of migration keywords was prepared in collaboration with social scientists Andreja Vezovnik and Veronika Bajt. It contains 95 lemmas with all their corresponding word forms prepared by Anka Supej.}. The speech selection is very straightforward. If any of our keywords appear in the transcription (observed as lemmas), we select the speech. Table~\ref{tab:table1} shows a general overview of the dataset magnitude, while Figure~\ref{figs:piechart} shows a pie chart of speech share selections from the set of keywords. The final dataset turned out to be quite balanced between the two classes, with 1519 of the speeches classified as ``left'', and 1455 classified as right.

\begin{table*}
\centering
\small
\begin{tabular}{l|r|r|r|r|r}
                                Dataset & \multicolumn{1}{c|}{\#Parties} & \multicolumn{1}{c|}{\#Speakers (Left/Right)} & \multicolumn{1}{c|}{\#Speeches (Left/Right)} & \multicolumn{1}{c|}{AWS} & \multicolumn{1}{c}{MSS} \\ \hline
SI ParlaMint                    & 12                             & 353 (114/48)                                 & 75122 (70.6\%/20.4\%)                        & 1244.1                   & 30.0                    \\ \hline
SI ParlaMint (MPs)              & 12                             & 166 (114/48)                                 & 31151 (55.9\%/40.9\%)                        & 1210.5                   & 125.5                   \\ \hline
SI Parlamint (MPs on migration) & 12                             & 153 (105/48)                                 & 2974 (51.1\%/48.9\%)                         & 854.7                    & 12.0                   
\end{tabular}
\vspace*{-1ex}
\caption{\textbf{General statistics of the Slovenian ParlaMint dataset and two of its subsets.} AWS - Average words per speech. MSS - Median speeches per speaker.}
\vspace*{-2ex}
\label{tab:table1}
\end{table*}

\section{Models and Explainability}
\label{modeling}

We followed two approaches to training machine learning models for text classification. The first is a classical bag of words feature extraction combined with a Linear SVM classifier~\cite{liu2010study}. The second approach is using the state-of-the-art Transformer language models~\cite{vaswani2017attention}. We applied several variations of both approaches, evaluated using 5-fold cross-validation, and compared the prediction accuracy. Finally, we selected the best model for each approach and trained it on the full dataset in order to apply explainability analysis.

For the classical approach, we trained a Linear SVM on three different bag-of-words types: unigram; unigram and bigram; unigram, bigram, and trigram. As the common practice, we used only lemmas which were already available in SI ParlaMint. We used a well-refined stopwords list for the Slovenian language to remove noise from the bag of words\footnote{https://github.com/stopwords-iso/stopwords-sl}, but we also used a minimum frequency of 5 appearances in the whole dataset, and a maximum frequency of a word appearing in 35\% of the speeches which proved effective in removing the procedural parts of speeches and improving the results. We optimized the model on each of the training folds in a nested 5-fold cross-validation and we evaluated the optimized model on the main testing folds, practically optimizing and evaluating five models of the same kind, reporting the average and its 95\% confidence interval. Finally, using the hyperparameters of the best-performing model, we trained one final model which we used for the explainability experiments.

Opposed to the common understanding of machine learning models, a Linear SVM classifier is not a black box that cannot be explained in its calculations. It creates a hyperplane that uses support vectors to maximize the distance between the two predicted classes (in our case ``left'' and ``right''). The weights that represent the Linear SVM equation hyperplane are the vector coordinates that are orthogonal to the hyperplane. So, their direction indicates the predicted class. The absolute size of the coefficients in relation to each other can then be used to determine feature importance for the data separation task. For our task, the minimum (most negative) weight coefficients correspond to words in the bag-of-words representation that led the model to classify a speech as a ``leftist'', while the maximum (most positive) weight coefficients correspond to words that led it to classify a speech as ``right''. Using this explainability logic, we derived an understanding of what makes the difference between a leftist and a rightist ideology's thought and word choice.

For language models, we used two types: the multilingual BERT and the SloBERTa. The multilingual BERT~\cite{DBLP:journals/corr/abs-1810-04805} was trained in 104 languages and is one of the first successful language models which serves as an appropriate baseline for fine-tuning the model to any kind of textual task. SloBERTa~\cite{ulvcar2021sloberta} uses the BERT architecture and it is monolingual, meaning it is trained purely on Slovenian data, making it more suitable for tasks where the model is applied specifically to Slovenian text, as is ours. Both pre-trained models were fine-tuned on the raw SI ParlaMint transcriptions using 30 epochs. The Learning rate of the BERT model was 5e-6, for the SloBERTa was 6e-6. As in the SVM case, we optimized the number of epochs and the learning rate using 5-fold nested cross-validation. For preprocessing, we used the standard respective tokenizers of BERT and SloBERTa. As both models are limited to 512 tokens, we applied an automatic truncation, using the first 512 words in each speech for training and predicting. Both models are limited to a maximum of 512 tokens, and speeches above that limit were truncated.

For explaining the deep learning models, we used a technique introduced \citet{NIPS2017_7062} called by SHAP. It is a powerful technique that uses classic equations from cooperative game theory to compute explanations of model predictions. Shapley values are feature importance values derived when a model is trained on all feature subset combinations, mathematically calculating the importance of each. Calculating Shapley values in this manner is computationally expensive, so what SHAP manages to do is derive these values by sampling approximations. Reading the SHAP results is very similar to reading the explainability analysis of the Linear SVM: more negative values refer to words leading the model to classify a speech as ``leftist'' and more positive values refer to the ``rightist'' words. There are two ways of presenting the token importance using Shapley values: its total Shapley in the entire dataset or its maximum/minimum recorded value. The issue with the first technique is that it prioritizes common words such as stopwords and should be generally avoided. We use the second technique, as it leads to more sensible results. A third variation of normalizing the Shapley sum according to the number of occurrences is also an option, yet we leave out this experimentation and plan it as our further work.

\section{Results and Discussion}
\label{results_discussion}

Table~\ref{tab:table2} shows the results of classifying speeches on the topic of migration as ``leftist'' or ``rightist'', depending on the political affiliation of the speaker in the Slovenian Parliament. The classical Tf-idf approach using unigrams, bigrams and trigrams showed the best results, with around 91\% accuracy. From the language models, the SloBERTa outperformed BERT yet with only 87\% accuracy. One reason behind the worse performance of the language models could be its inability to operate on very long sequences, as more than 60\% of the speeches in our dataset contain more than 512 tokens (maximum sequence for BERT).

\begin{table}[]
\small
\centering
\begin{tabular}{l|r}
Model                                      & Accuracy            \\ \hline
Random Baseline                            & 0.511               \\ \hline
Tf-Idf + Lin. SVM (1-grams)                & $0.866 \pm 0.01 $        \\
Tf-Idf + Lin. SVM (1, 2-grams)             & $0.903 \pm 0.01 $        \\
\textbf{Tf-Idf + Lin. SVM (1, 2, 3-grams)} & \bm{$0.913 \pm 0.02$} \\ \hline
Multilingual BERT                          & $0.819 \pm 0.01$         \\
\textbf{SloBERTa}                          & \bm{$0.877 \pm 0.03$}
\end{tabular}
\vspace*{-2ex}
\caption{\textbf{Classification scores on predicting political leanings based on speech transcriptions.}}
\vspace*{-3ex}
\label{tab:table2}
\end{table}

\begin{figure*}[t!]
\centering
\includegraphics[width=0.85\textwidth]{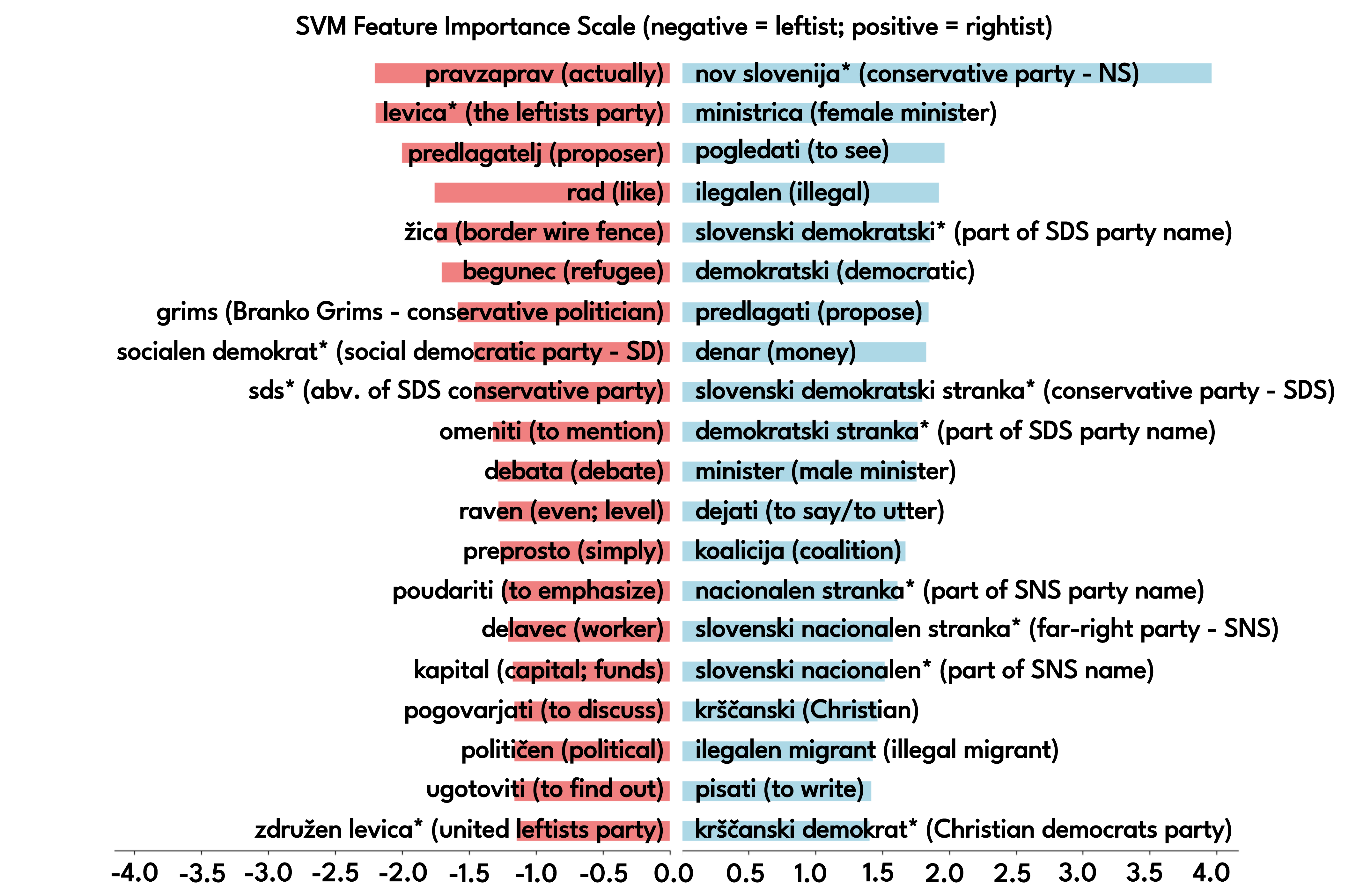}
\vspace*{-2ex}
\caption{\textbf{SVM feature importance.} The left (red) side contains phrases that have the highest significance for the model predicting a speech as ``leftist''. The same explanation is for the right (blue) side for the ``rightist'' speeches. Phrases marked with \texttt{*} refer to party names.}
\vspace*{-2ex}
\label{figs:svm_featureimportance}       
\end{figure*}

\begin{figure*}[t!]
\centering
\includegraphics[width=0.99\textwidth]{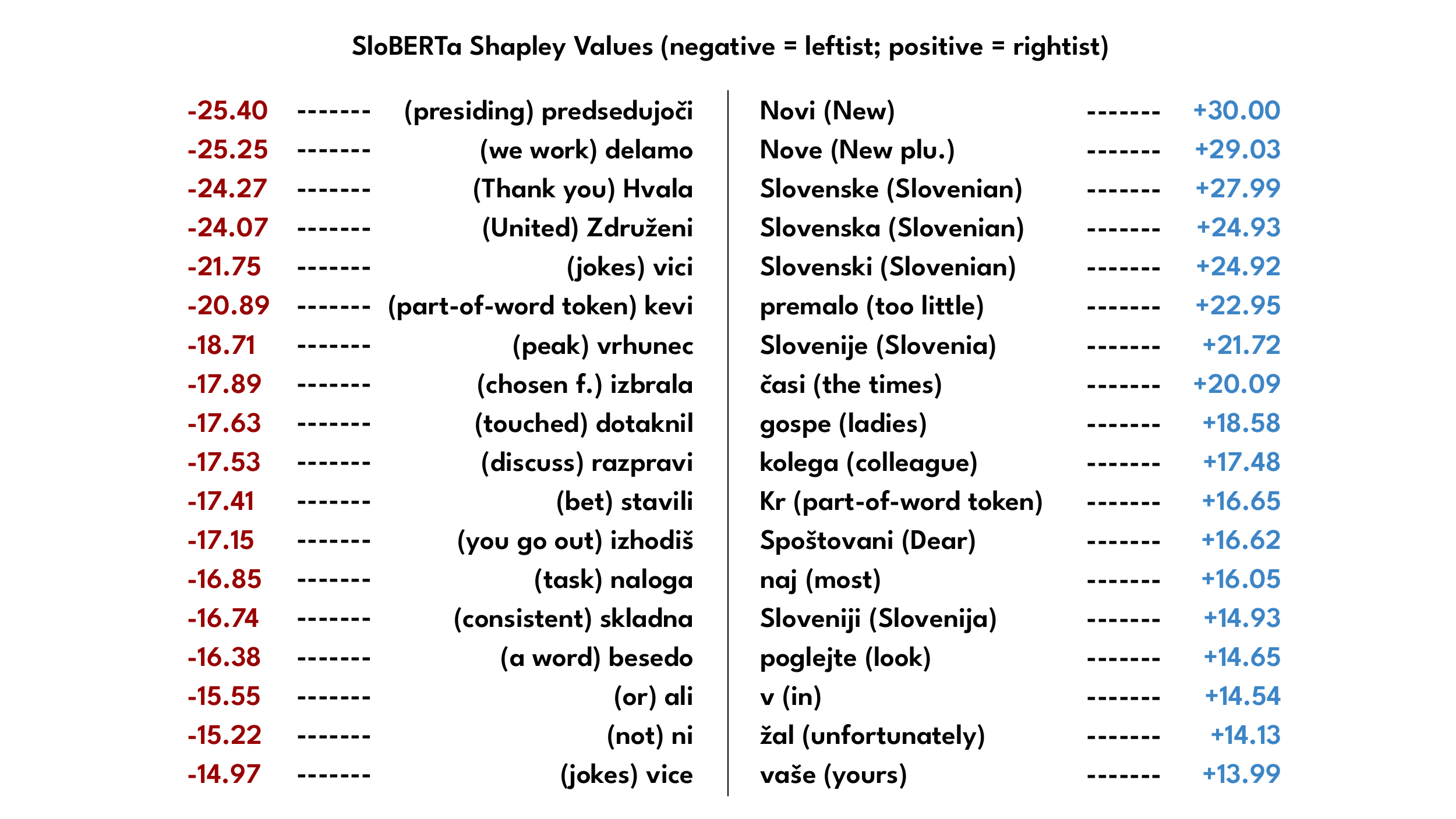}
\vspace*{-2ex}
\caption{\textbf{Maximum and minimum Shapley values for the SloBERTa model.} Left/right (red/blue) side shows the words that had the highest signal in predicting a ``leftist/rightist'' speech.}
\vspace*{-2ex}
\label{figs:sloberta_featureimportance}       
\end{figure*}

The models showed that predicting parliamentary political leaning from speech transcriptions is possible. Since statistical models can differentiate ``leftist'' and ``rightist'' speeches, we investigated if their explanation through feature importance makes sense from a political linguistics perspective. Figures~\ref{figs:svm_featureimportance} and~\ref{figs:sloberta_featureimportance} show the results of the feature importance analysis. Here, we can notice some obvious similarities and differences between the classical and the language model approach. The first difference is that the SVM feature importance can catch not only tokens (words) but also bigrams and trigrams, which can be useful for a better linguistic context. Since it uses lowercase lemmas, it mitigates duplicates in different word forms, yet it can potentially lead to ambiguities. The most important difference is that the SVM feature importances are the general values for the whole dataset, while the Shapley values are detected maximums for a particular comment, which does not mean that these words have the same negative (leftist) or positive (rightist) effect for all samples.

Both model interpretations show that right-wing parliamentarians' motif is to mention the country name (Slovenia) and its forms. The SVM interpretation also reveals that the rightists emphasize their party names. Regarding migration, their motif is to focus on the illegal aspect of the migration, with the SVM models catching the phrase ``illegal migrant'' as something that almost exclusively right-wing politicians use. The leftists tend to use the word ``begunec'' (angl. refugee) instead of ``migrant'', as the meaning behind the former refers to someone who is fleeing. 
They tend to use words such as ``united'' and ``debate'' and they often times refer to their party opponents by the party abbreviations instead of the full name. One interesting case is that the SVM manages to recognize the leftists emphasizing the word ``grims''. Grims is the surname of the right-wing politician Branko Grims who is strongly against immigrants in Slovenia and Europe and was the most active politician in the parliament on the topic.

Although much more can be drawn from the feature importance, this is where our analysis stops. The model interpretability through the list of words that had the most meaningful impact on the classification opens up a great possibility for further qualitative work. Researchers studying political linguistics could observe the usages of these lists and find patterns in the connotation of the words used. They could confirm or debunk ideological concepts used by the different sides of the political spectrum. Or, they could analyze if these words and phrases actually contain the primary message of a sentence or if their role is more on the stylistic side of political speeches.

\section{Conclusion}
\label{conclusion}

Explaining the decision-making of machine learning models (known as XAI) can be a great tool in interdisciplinary research. When the goal is not just to classify, but to understand patterns that appear across classification groups, XAI can complement qualitative research.

In this work, we used XAI to bridge the gap between computational and political linguistics. We developed classical machine learning as well as deep learning language models that can classify parliamentary speeches as ``leftist'' and ``rightist'' for the topic of migration. We applied our approach to the Slovenian parliament using the ParlaMint dataset with data from 2014 to 2020. With both approaches showing great predictive success, we applied methods (such as calculating the Shapley values) that can explain the decisions of our models and show which words and phrases differentiate ``leftist'' from ``rightist'' speeches and vice versa. While left-leaning parliamentarians use concepts such as ``unity'' and ``debate'', the right-leaning parliamentarians put more emphasis on the national symbols (mentioning Slovenia) and their party names.

We leave an opening for further work in multiple directions. One is the improvement of the interpretability methods for models that work with text. Different feature importance techniques could be applied and a comparative study could help us understand the strengths and weaknesses of the methods. Another direction is exploring the ways the model explanations can be used in interdisciplinary research. How to use the words that give the most value to models' predictions in a qualitative continuation of the work. Or, how to understand and communicate model flaws with experts from the social sciences in order to improve them.

\paragraph{Acknowledgements}
The work was supported by the Slovenian Research Agency research projects and programmes P6-0436: Digital Humanities: resources, tools and methods (2022–2027), P6-0280: Economic, Social and Environmental History of Slovenia (2022–2027), P2-0103: Knowledge Technologies  (2022–2027), J6-2581: Computer-assisted multilingual news discourse analysis with contextual embeddings  (2020-2023) and J5-3102: Hate speech in contemporary conceptualizations of nationalism, racism, gender and migration (2021-2024), as well as the DARIAH-SI research infrastructure (2022-2027). We also acknowledge the financial support from the RobaCOFI project, which has indirectly received funding from the European Union’s Horizon 2020 research and innovation action programme via the AI4Media Open Call \#1 issued and executed under the AI4Media project (Grant Agreement no. 951911). We also thank Andreja Vezovnik and Veronika Bajt for their help in preparing the migration keywords.

\paragraph{Code}
The source code is publicly available at: \url{github.com/boevkoski/xaicl_parliaments}


\bibliographystyle{ltc23}
\bibliography{bibliography} 

\end{document}